\documentclass[review]{elsarticle}

\usepackage{lineno,hyperref}
\usepackage[english]{babel}
\usepackage[utf8x]{inputenc}
\usepackage{amsmath,amssymb}
\usepackage{bm}
\usepackage{graphicx}
\graphicspath{{Figures/}}
\usepackage{siunitx}
\usepackage{booktabs}
\usepackage{mathtools}

\usepackage{caption}
\usepackage{numprint}
\usepackage{multirow}
\usepackage[ruled,linesnumbered]{algorithm2e}
\usepackage{chngcntr}
\usepackage{parskip}
\usepackage{mathtools}
\usepackage{makecell}

\modulolinenumbers[5]
\usepackage[top=30mm, bottom=25mm, right=30mm, left=30mm]{geometry}

\begin{document}

\begin{frontmatter}

\title{Physically-informed change-point kernels for structural dynamics}

\author[Address1]{D.J. Pitchforth}

\author[Address1]{M.R. Jones}
\author[Address1]{S.J. Gibson}
\author[Address1]{E.J. Cross}

\address[Address1]{Dynamics Research Group, Department of Mechanical Engineering, University of Sheffield, Mappin Street, Sheffield S1 3JD, United Kingdom}

\begin{abstract}
The relative balance between physics and data within any physics-informed machine learner is an important modelling consideration to ensure that the benefits of both physics and data-based approaches are maximised. An over reliance on physical knowledge can be detrimental, particularly when the physics-based component of a model may not accurately represent the true underlying system. An underutilisation of physical knowledge potentially wastes a valuable resource, along with benefits in model interpretability and reduced demand for expensive data collection. Achieving an optimal physics-data balance is a challenging aspect of model design, particularly if the level varies through time; for example, one might have a physical approximation, only valid within particular regimes, or a physical phenomenon may be known to only occur when given conditions are met (e.g.\ at high temperatures). 

This paper develops novel, physically-informed, change-point kernels for Gaussian processes, capable of dynamically varying the reliance upon available physical knowledge. A high level of control is granted to a user, allowing for the definition of conditions in which they believe a phenomena should occur and the rate at which the knowledge should be phased in and out of a model. In circumstances where users may be less certain, the switching reliance upon physical knowledge may be automatically learned and recovered from the model in an interpretable and intuitive manner. Variation of the modelled noise based on the physical phenomena occurring is also implemented to provide a more representative capture of uncertainty alongside predictions. The capabilities of the new kernel structures are explored through the use of two engineering case studies: the directional wind loading of a cable-stayed bridge and the prediction of aircraft wing strain during in-flight manoeuvring.
\end{abstract}

\end{frontmatter}


\newpage
\section{Introduction}

Engineering knowledge and our understanding of how physical systems behave is a valuable resource. Physics-informed Machine Learning (PIML) aims to utilise this resource to enhance the capabilities of data driven models, by improving performance, and providing insight into their results. Although highly flexible, and capable of capturing complex relationships, purely data-based learners can only expect to predict reliably within the realms of previously observed conditions. For many engineering structures, particularly those in harsh environments, the possible set of observed conditions is too large to feasibly measure completely. Through leveraging physical knowledge, PIML offers a potential solution to reduce the demand for data collection to a more attainable level.

Three of the major decisions within a PIML model are the type of physics to be included, the selection of a data-based component, and how they are integrated together to create a single model. The latter of these decisions is the focus of this paper and is a key factor effecting the relative balance between physics and data within the model. Intuitively, the higher the extent to which a system or process is understood, the more one might wish to rely upon physical knowledge. Earlier work of the authors \cite{Cross2024SpectrumGP} discusses the relationship between how the structure of Gaussian process models affects this balance and how it may change across modelling tasks.

This work looks toward how one might be able to tune the relative reliance on physics and data within physics-informed Gaussian Process (GP) models. A GP is a flexible, non-parametric, Bayesian technique, adept within a variety of engineering regression problems, including crack growth \cite{Pfingstl2022GPmeanCrack,An2015GPmeanCrack}, tool wear \cite{Toth2023GPmeanTool,Kong2018GPmeanTool} and modelling of wind turbine power curves \cite{Rogers2020powerWT,Bull2022powerWT}. For conciseness, an introduction to GP regression theory is not presented here, with interested readers encouraged to consult \cite{GPRasmussen}. The covariance function (kernel) of a Gaussian process is responsible for the family of functions from which predictions may be drawn, with commonly used kernels enforcing properties such as smoothly varying functions, periodicity and localised behaviours \cite{GPRasmussen,GPTutorial}. Through careful design of the kernel, it is possible to mimic physically desirable behaviours within drawn functions; such examples include the representation of a physical process \cite{Cross2022SDOFkernel,Kok2018MagFieldSLAM} or the enforcement of axial and rotational symmetries \cite{Noack2022AdvKernelDesign,HAYWOODALEXANDER2021StructuredML,Klus2021AntiSymmetricKernel,Grisafi2018SymmetryML}. Here, a combination of structured, physics-informed kernels and flexible, more general purpose, kernels are used in combination. 

A key theme of this paper is framing physics-informed machine learning within a switching model context, where a switching between model components determines the relative reliance upon physical knowledge and measured data. There are many existing architectures that one might employ for this task, including Mixtures of Experts (MOEs) \cite{Baldacchino2016BayesMOE,Baldacchino2017BayesMOERobust}, Markov-switching models \cite{Kim1994MarkovSwitch}, Treed Gaussian Processes (TGPs) \cite{Worden2018TreeGP} and regime-switching cointegration models \cite{Shi2018SwitchCoint}. Here, due a natural fit within a Gaussian process regression framework, a change-point structure \cite{wilson2013changept,wilson2014thesis} is adopted. This incorporates model switching directly within the design of the kernel, allowing for ease of integration within a typical Gaussian process regression pipeline. Access to the contributions of individual kernel components and information about when, and how quickly, switches occur are also able to provide valuable insight into model operation.



Intermittent physical phenomena occur in many engineering structures, and depending on their rarity, a scarcity of data may make them challenging to model using purely data-based techniques \cite{Wan2020IntermittentLeak,Belenky2012RareShipEvent,Xue2023FreakWave}. This paper introduces how one might phase in physical knowledge to aid model predictions when such phenomena occur. Firstly, a case study is presented of wind loading of the Tamar bridge \cite{Koo2013Tamar}. A section of a dataset measured on the structure is used to investigate lift forces produced by high speed winds. Importantly, these lift forces only occur when winds blow across the bridge (perpendicular to the bridge length), causing a dependency of this relationship on wind direction. The implementation of physics-informed change-point kernels is used to capture the dynamic reliance on lift force, outperforming a purely data-based approach, and providing insight in to how this relationship changes. We additionally show how change-point kernels can be combined with a heteroscedastic likelihood model to capture input dependent observation noise - for the Tamar bridge, this behaviour arises as noise that grows with wind speed, where we see wider variation on deck acceleration values as the wind speed increases. For the second case study, a familiar example of engineering knowledge, the oscillation of a cantilever beam, is embedded within a Gaussian process kernel to aid predictions of aircraft wing strain in-flight. This is phased in and out of the model using a change-point structure, utilising the inputs of a pilot to change the relative explanatory power of kernel components.




%
\section{Change-point kernels}
The selection of a kernel within a Gaussian process is an important modelling decision, determining the type of functions used to make predictions and, as a result, the model's performance. However, there are many instances where the selection of a single `best' kernel for a modelling task might be challenging, for example, a system under changing conditions or the introduction of new behaviours over time. In this work, we build upon the change-point kernel \cite{wilson2013changept,wilson2014thesis}, which controls the switching between two kernels $K_1(X,X')$ and $K_2(X,X')$ through the use of the sigmoid function:
\begin{equation}
\sigma(x) = \frac{1}{1+e^{-a(x-x_0)}}
\end{equation}

\noindent where $a$ is a gradient term, responsible for how quickly the function switches from 0 to 1, and $x_0$ is the switching location. The use of a sigmoid function allows the gradual transition between multiple kernels without the introduction of a discontinuity. A useful property of sigmoid functions is that $\sigma(x)+\sigma(-x)=1$ allowing a pair of opposing sigmoids to phase between the use of two kernels. A sigmoid with a negative gradient term may be used to phase out the use of a particular kernel, whilst a sigmoid with the equivalent positive gradient may be used to phase in the use of a new kernel. The covariance function for a sigmoid is expressed:
\begin{equation}
K_{\sigma}(X,X') = \sigma(X)\sigma(X')^T
\end{equation}

Given that products and sums of kernels are also valid kernels, a pair of opposing sigmoid kernels may be used to switch between the use of two kernels $K_1(X,X')$ and $K_2(X,X')$. This is referred to as the change-point kernel \cite{wilson2013changept}:
\begin{equation}
  K(X,X') = K_{\sigma}(X,X')K_1(X,X') + K_{\sigma}(-X,-X')K_2(X,X')
\end{equation}

where additionally to the hyperparameters of the kernels $K_1(X,X')$ and $K_2(X,X')$, a gradient term, $a$ is introduced to control the direction and speed of the switch and the location, $x_0$ determines where the switch happens.

\begin{figure}[ht]
\centering
\includegraphics[width=1\textwidth]{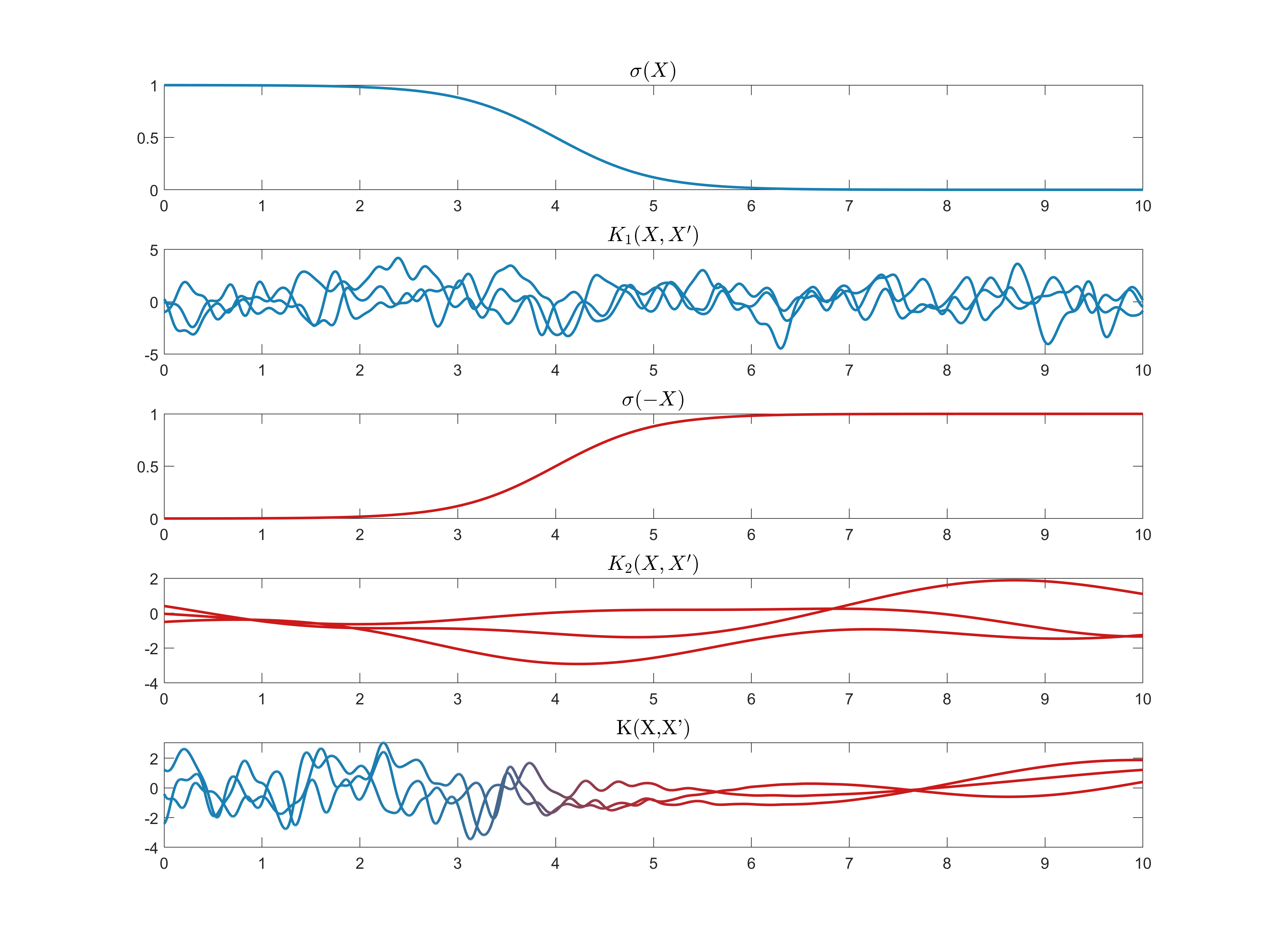}
\caption{Construction of a change-point kernel with $a=2$ and $x_0=4$, transitioning between Squared Exponential kernels with short and long lengthscales. The sigmoid functions,  $\sigma(X)$ and $\sigma(-X)$, draws from each component kernel, $K_1(X,X')$ and $K_2(X,X')$, along with draws from the combined kernel $K(X,X') = K_{\sigma}(X,X')K_1(X,X') + K_{\sigma}(-X,-X')K_2(X,X')$ are shown. The colour gradient of the plot reflects the relative weighting of each kernel; blue for $K_1(X,X')$, and red for $K_2(X,X')$.}
\label{fig:ChangePt_SE_draws_subplot}
\end{figure}

To highlight the operation of a change-point kernel, consider a case where one may wish to transition from a process that varies quickly, to one varying more slowly. This can be achieved through a combination of short and long lengthscale Squared Exponential (SE) kernels. Draws from a SE change-point kernel are shown in Figure \ref{fig:ChangePt_SE_draws_subplot} for illustration.

\subsection{Physics-informed change-point kernels}

The framework of change-point kernels provides an effective means to vary the reliance between multiple kernels within a GP. To exploit this within a PIML setting, the integration of physics-informed kernels and flexible `data-based' kernels is proposed here, alongside the option to switch on an input variable separate from those used in the main GP. Here, a pair of sigmoid kernels $K_{\sigma}(Z,Z')$ and $K_{\sigma}(-Z,-Z')$ are used to control the explanatory power of the physics-informed kernel within the model. This leads to a kernel structure of:

\begin{equation}
  K(Z,Z',X,X') = \underbrace{K_{\sigma}(Z,Z')K_{Phy}(X,X') }_{\begin{matrix}{\text{Physics-informed kernel}}\end{matrix}} + \underbrace{K_{\sigma}(-Z,-Z')K_{Data}(X,X') }_{\begin{matrix}{\text{Flexible kernel}}\end{matrix}}
\end{equation}

where the newly introduced $Z$ is the input to the sigmoid kernel, used to control the switching between kernels $K_{Phy}(X,X')$ and $K_{Data}(X,X')$. An effective choice of $Z$ relates closely to how a physical relationship might change with a variable. For example, temperature, humidity, excitation level, or measures of turbulence can all effect the extent to which one might want to rely on a given piece of physical knowledge.

There are many reasons why one might want to change the relative reliance on physical knowledge within a model, an important one of which is the changing validity of a physics-based model. With any physical model, and particularly so with simple ones, assumptions must be made in order to represent the system of interest. The extent to which these assumptions hold effects the performance of the constructed model and care should be taken not to trust the results of models constructed upon invalid assumptions. Assigning a fixed degree of trust within a physical model component may be challenging when a model is required to operate over a range of conditions and allowing this to vary is therefore highly desirable.

The presence of regime-switching and localised behaviours provide alternative motivation to vary the reliance on physical knowledge. If a phenomena is known to occur in specific conditions, for example the dependency of vortex shedding on flow speed \cite{Sarpkaya2010WaveForces}, one might wish to phase its occurrence in and out of a model. To investigate a use-case of this form, a case study of wind loading on the Tamar bridge will now be presented.




%
\begin{figure}[ht]
\centering
\includegraphics[width=1\textwidth]{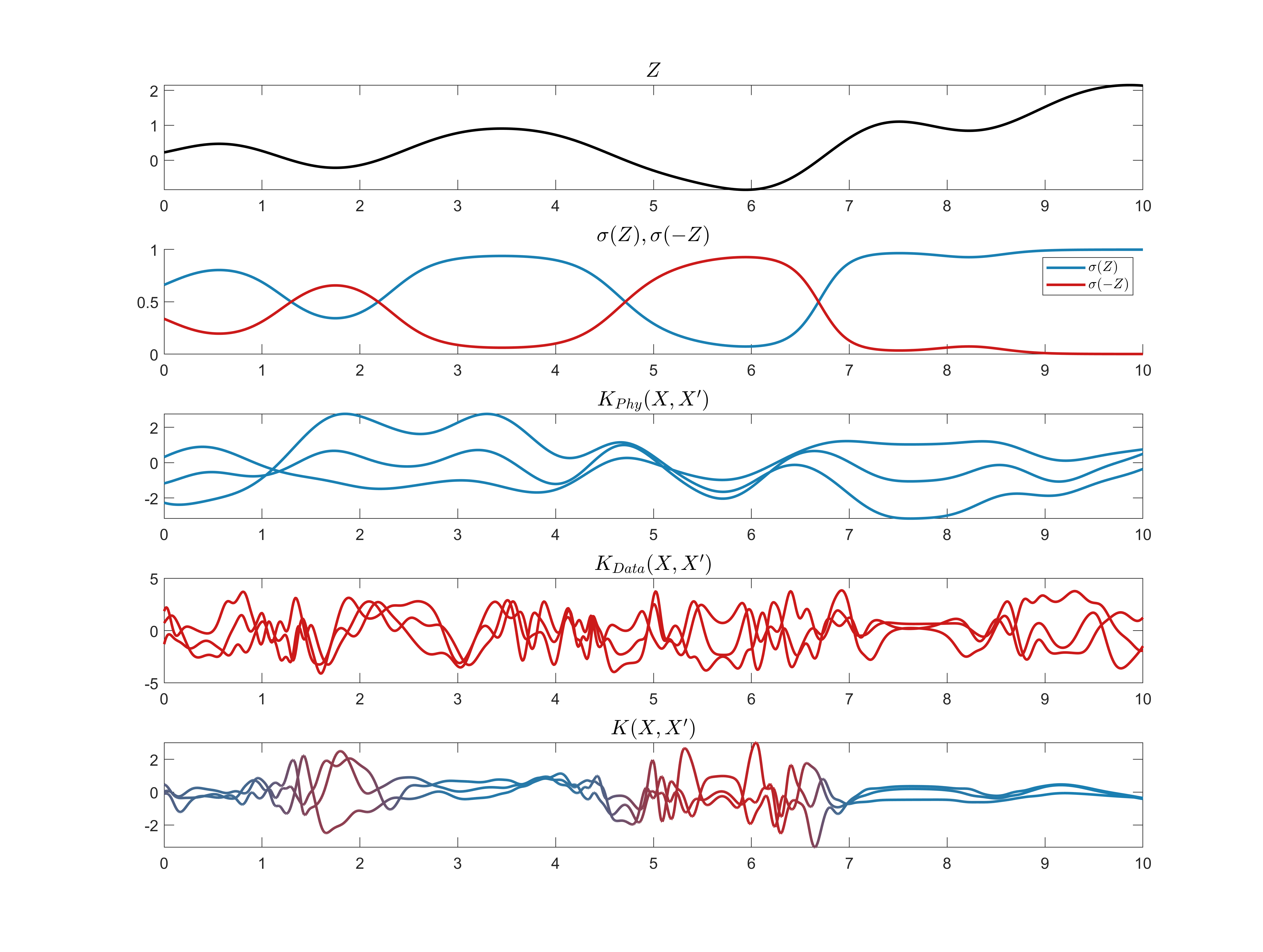}
\caption{Construction of a physics-informed change-point kernel, where the switching between a physics-informed kernel $K_{Phy}$ and a flexible kernel $K_{Data}$ is controlled by an external variable $Z$. Here $K_{Phy}(X,X')$ is a linear kernel, acting on an input space of $X = [\bm{x1},  \bm{x2}]$, representing knowledge of a linear process. This linear relationship is assumed not to hold for lower values of $Z$. The colour gradient of the plot reflects the relative weighting of each kernel; blue for $K_{Phy}(X,X')$, and red for $K_{Data}(X,X')$.}
\label{fig:ChangePt_Tau_SE_draws_subplot_with_tau}
\end{figure}

\section{Case study A: Directional wind loading of the Tamar bridge}

The Tamar Suspension Bridge is located in the south-west of the UK and was the feature of a monitoring campaign led by the Vibration Engineering Section (VES) at the University of Sheffield. The bridge is 643m long, with two towers, each 73m in height; it also forms part of a major connection to the city of Plymouth. For additional details on both the bridge and monitoring campaign, the reader is directed towards \cite{LizzyThesis} and \cite{Koo2013Tamar}.

\begin{figure}[ht!]
  \centering
  \includegraphics[width=0.9\textwidth]{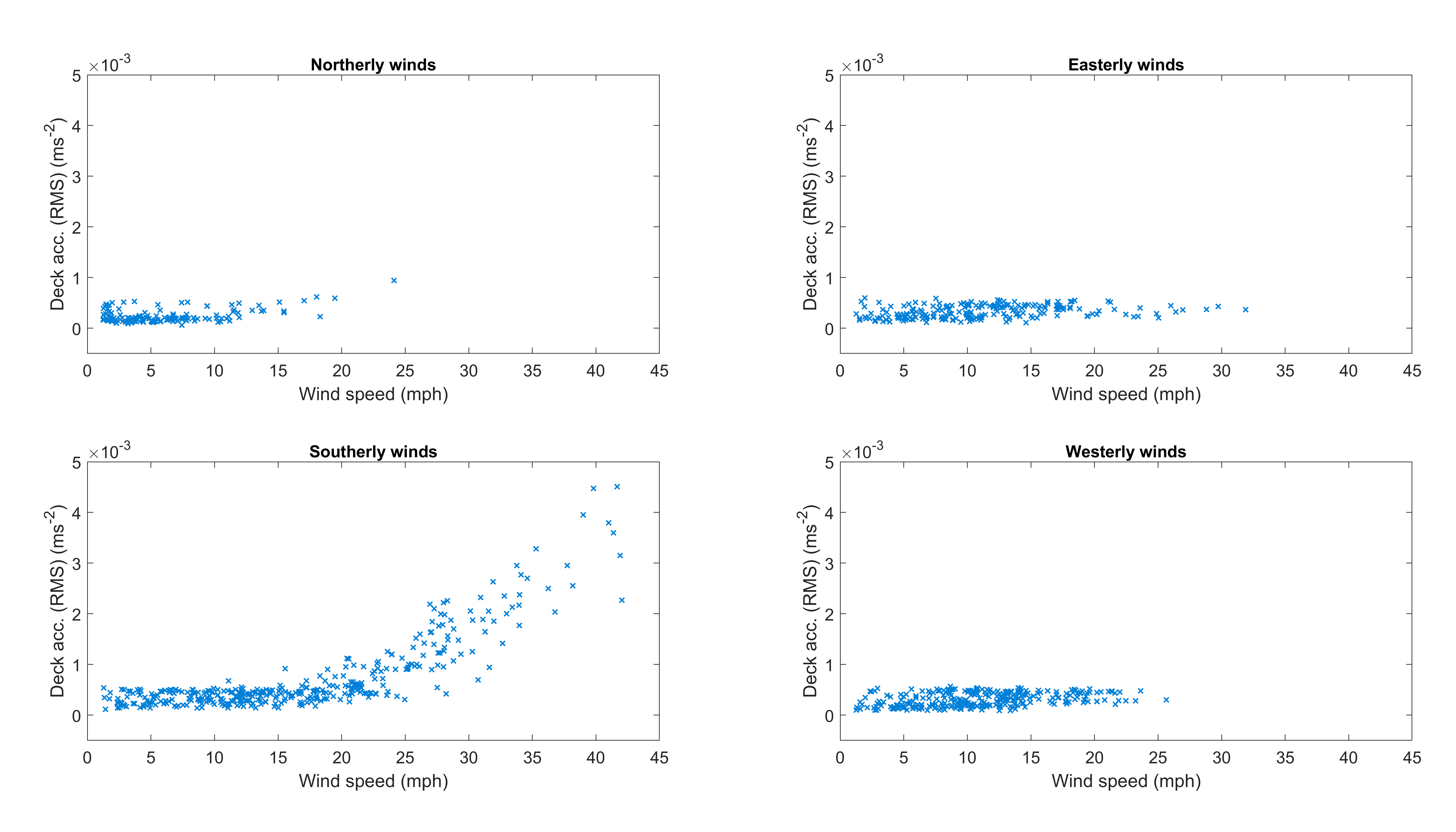}
  \caption{Wind speed vs vertical deck acceleration for a section of the Tamar bridge dataset. Winds are separated by direction, into northerly winds ($0 \pm 20^\circ $), easterly winds ($90 \pm 20^\circ $), southerly winds ($180 \pm 20^\circ $) and westerly winds ($270 \pm 20^\circ $). The northerly and southerly winds blow sideways across the bridge and induce much higher deck accelerations, particularly at high wind speeds.}
  \label{fig:NESW_Subplots_data_only}
\end{figure}

Available data from the bridge spans a three-year period from 2007 to 2011, with access to measurements from accelerometers, strain gauges, anemometers, temperatures, humidity and traffic levels. Here, only a small section of this dataset (2500 points) is used from the summer of 2008, focussing on measurements of wind speed, wind direction and vertical deck acceleration. The aim of this case study is to investigate the presence of changing physical behaviours from a full scale engineering structure. The example chosen here is the dependency of response to wind load on wind direction. Orientated in the east-west direction, the bridge is subject to higher deck accelerations when winds blow perpendicular to this at high speeds \cite{LizzyThesis}. This is shown in plots of wind speed vs deck acceleration, separated by wind direction in Figure \ref{fig:NESW_Subplots_data_only}.

To represent the changing behaviour present within the deck response $y(U,\theta)$, the problem is formulated as:
\begin{equation}\label{eq:tamar_obs_model}
y(U,\theta) = f_1(U) + f_2(U,\theta) + \epsilon
\end{equation}

where $U$ is wind speed, $\theta$ is wind angle, $f_1(U)$ is the deck response to low-speed winds, $f_2(U,\theta)$ is the deck response to high-speed lateral winds and $\epsilon$ is noise. The cause of this directional relationship may be explained as lift force, incited by a pressure difference as wind flows over the bridge deck \cite{Kerenyi2009BridgeLoads}:
\begin{equation}
  L = \frac{1}{2}\rho C_L A U^2
\end{equation}

\noindent where $\rho$ is fluid density, $C_L$ is lift coefficient, $A$ is projected area and $U$ is wind speed. The quadratic term in this expression aligns with the observations in Figure \ref{fig:NESW_Subplots_data_only}, particularly for southerly winds. To incorporate this knowledge within our model, a second order polynomial kernel is used to represent lift force:
\begin{equation}
  K_{Lift}(U,U') = \sigma_L^2 (UU^T+c)^2
\end{equation}

\noindent where $\sigma^2_L$ scales the variance and $c$ allows the introduction of lower order terms (linears and constants). Draws from this kernel will be quadratic, enforcing desirable structure within the predictions of the model. Using this kernel alone would not be sensible as this quadratic behaviour is clearly not visible for easterly and westerly winds. Instead, it is integrated within a larger change-point kernel structure:
\begin{equation}
  \begin{split}
  K(\theta,\theta ',U,U') & = \underbrace{K_{\sigma}(\cos(2\theta),\cos(2\theta'))K_{\sigma}(U,U')K_{Lift}(U,U') }_{\begin{matrix}{\text{Introduce lift force at high N/S winds...}}\end{matrix}} \\
  & + \underbrace{K_{\sigma}(-\cos(2\theta),-\cos(2\theta'))K_{\sigma}(-U,-U')K_{SE}(U,U') }_{\begin{matrix}{\text{... otherwise, use a Squared Exponential kernel.}}\end{matrix}}
\end{split}
\label{eq:TamarKernel}
\end{equation}

\noindent where $\theta$ is wind angle, $U$ is wind speed, $K_{\sigma}(\cos(2\theta),\cos(2\theta'))$ is a sigmoid kernel for wind direction, $K_{\sigma}(U,U')$ is a sigmoid kernel for wind speed, $K_{Lift}(U,U')$ is the polynomial lift kernel and $K_{SE}(U,U')$ is a Squared Exponential (SE) kernel. The aim of this structure is to allow the knowledge of lift force to be utilised only for higher speed northerly and southerly winds. The term $\cos(2\theta)$ is used a measure of how well a wind direction aligns with north or south.

\subsection{Results}
\vspace{-0.1cm}
To test the performance of models, a training set of 500 randomly selected points were used from the 2500 datapoint section of the dataset. The remaining unseen data was used as the test set. Models were optimised inline with standard GP practice \cite{GPRasmussen}, using the Negative Log Marginal Likelihood (NLML) as a cost function.

\begin{figure}[ht!]
  \centering
  \includegraphics[width=0.9\textwidth]{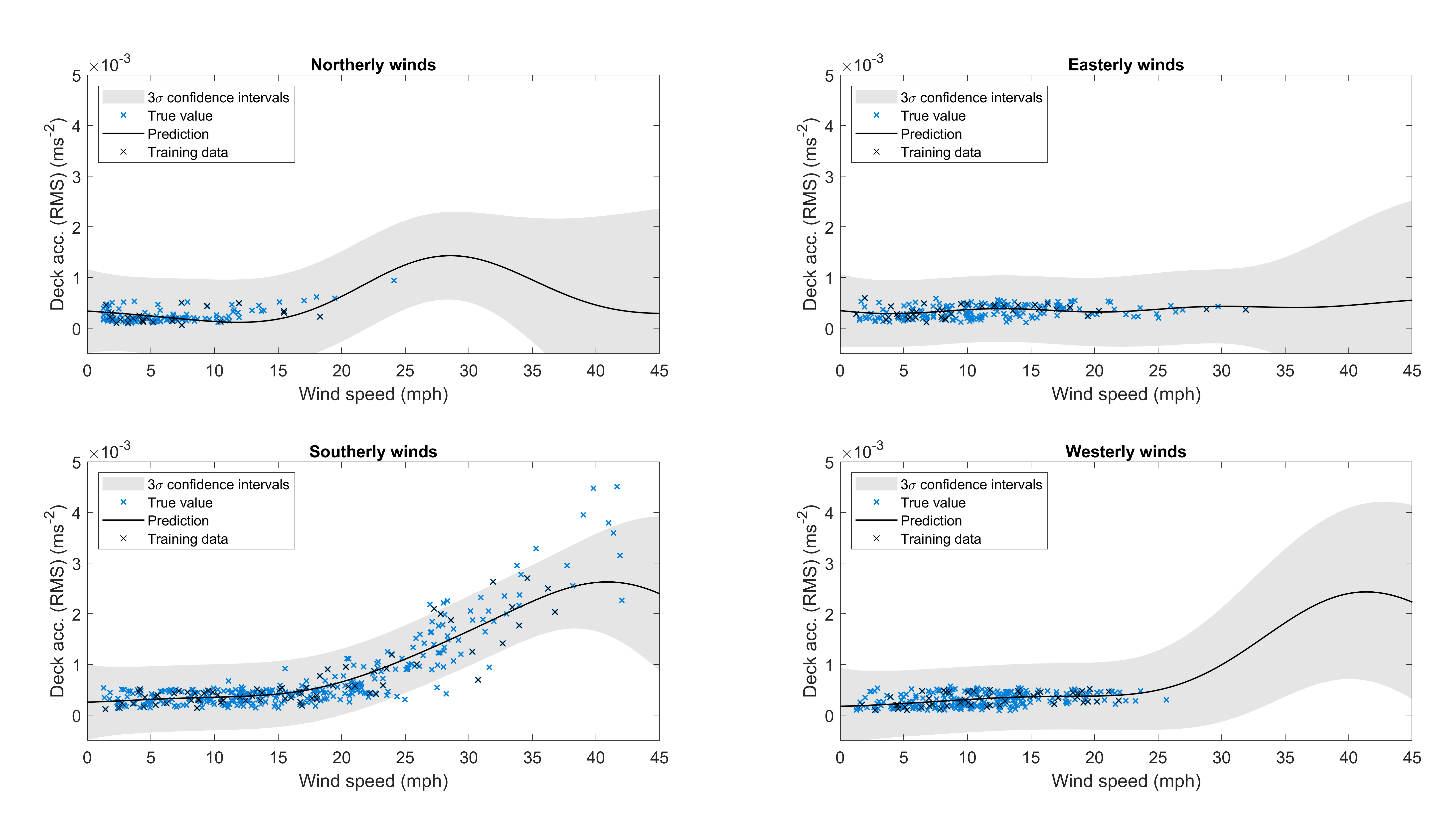}
  \caption{Predictions of vertical deck acceleration on the Tamar bridge, separated by wind direction, using a Gaussian process with a Squared Exponential kernel $K_{SE}(U,U',\theta,\theta')$. The model was shown a random scatter of 500 datapoints from a 2500 point dataset during training.}
  \label{fig:NESW_Subplots_SE_Kernel}

\vspace{0.2cm}
%
  \centering
  \includegraphics[width=0.9\textwidth]{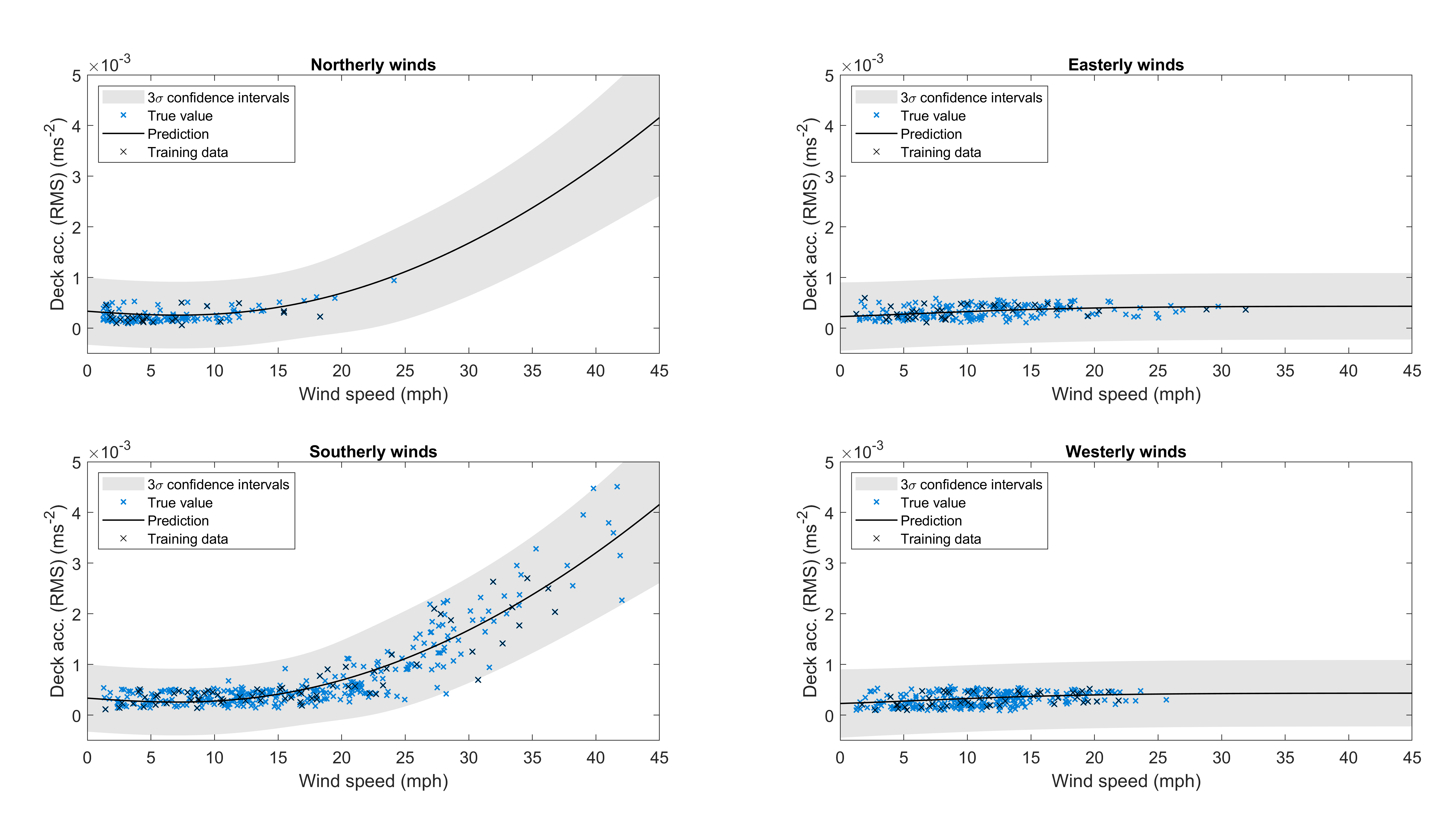}
  \caption{Predictions of vertical deck acceleration on the Tamar bridge, separated by wind direction, using a Gaussian process with a physics-informed change-point kernel. The model was shown the same random scatter of 500 datapoints from a 2500 point dataset during training as the SE kernel in Figure \ref{fig:NESW_Subplots_SE_Kernel}.}
  \label{fig:NESW_Subplots_Changepoint_Kernel}
\end{figure}

To allow the comparison of the proposed physics-informed change-points model with a benchmark case, a purely data-based model was also investigated. This took the form of a GP with a SE kernel, $K_{SE}(U,U',\theta,\theta')$, given access to the same model inputs of wind speed and wind direction. Predictions for this model are shown in Figure \ref{fig:NESW_Subplots_SE_Kernel}, with predictions from the change-point model shown in Figure \ref{fig:NESW_Subplots_Changepoint_Kernel}.

%

The first observation from the results is that the change-point model was better able to cope with the changing behaviours present within the dataset than the purely data-based approach. We can quantify this improvement with two error metrics; the normalised mean squared error (NMSE) and the mean standardised log loss (MSLL). Whilst the NMSE quantifies the deviation of the expectation of our predictions from the true value, the MSLL provides a measure of both mean predictive performance, in addition to how well the variation of the data has been captured. The MSLL is defined as,

\begin{equation}\label{eq:MSLL}
  MSLL = \frac{1}{N}\sum\{-\log p(\mathbf{y}_{\star}| X,\mathbf{y},X_{\star}) + \log p(\mathbf{y}_{\star};\mathbb{E}(\mathbf{y}), \mathbb{V}(\mathbf{y}))\},
\end{equation}

where the training set is denoted $\{X,\mathbf{y}\}$, test inputs $X_{\star}$ and test set observations $\mathbf{y}_\star$. As this metric is a measure of the negative log probability of the model, a lower value is indicative of a better fit.

Across the full time series, the change-point kernel achieved an NMSE of $24.25\%$ and a MSLL of -0.711, compared with the SE kernels' NMSE of $29.74\%$ NMSE and MSLL of -0.699. The introduction of the quadratic lift force for higher speed northerly and southerly winds can be seen, whilst importantly, it is not introduced in to the easterly or westerly winds. For the SE kernel however, some difficulty can been seen in distinguishing between behaviours. For example, the prediction of deck acceleration for westerly winds was pulled upwards by the presence of high speed southerly training data. Although these points do occupy different places in the input space $[U, \theta]$, the SE kernel could not effectively separate these relationships. A suspected reason for this is an over estimation for the lengthscale acting on wind angle, preventing predictions varying quickly enough w.r.t this variable.

An advantage of many PIML methods, also seen here, is an ability to use physical knowledge to assist with extrapolation. Beyond the highest northerly and southerly wind speeds observed, the change-point model is able to rely on the non-stationary lift kernel to predict further from observed data. For the SE kernel however, predictions tended back towards the prior mean of zero. For the section of the dataset used for this work, there was a lack of high speed northerly wind data, making predictions in this region challenging. In the UK, the prevailing wind direction is typically south-westerly, causing an increased likelihood to encounter higher wind speeds in this direction \cite{Lapworth2008PrevWindUK}. This is further exaggerated near the south-west coast, where the Tamar bridge is located. The rarity of higher speed northerly winds does not make a models' performance during their occurrence any less important and an effective model should be able to cope with reduced data in this region. There are many reasons why data for particular event may be rarer to measure and the use of physical knowledge is one way to try alleviate this.

A point of discussion relevant to kernel design is the number of hyperparameters within the model. Typically, the construction of more complex kernels comes at the expense of the introduction of additional hyperparameters, therefore increasing computational demands for optimisation. Here, the number of hyperparameters grows from three for the SE kernel, to eleven for the physics-informed change-point kernel leading to an increase in computation time by a factor of $1.7$. To alleviate this, several of the introduced hyperparameters were bounded during the optimisation. For example, the switching location of the wind speed sigmoid $x_{w0}$ was bounded between 5 and 30 mph. The use of physical knowledge to reduce optimiser search space is one of the advantages of hyperparameters with physical meaning.

\begin{figure}[ht]
  \centering
  \includegraphics[width=0.9\textwidth]{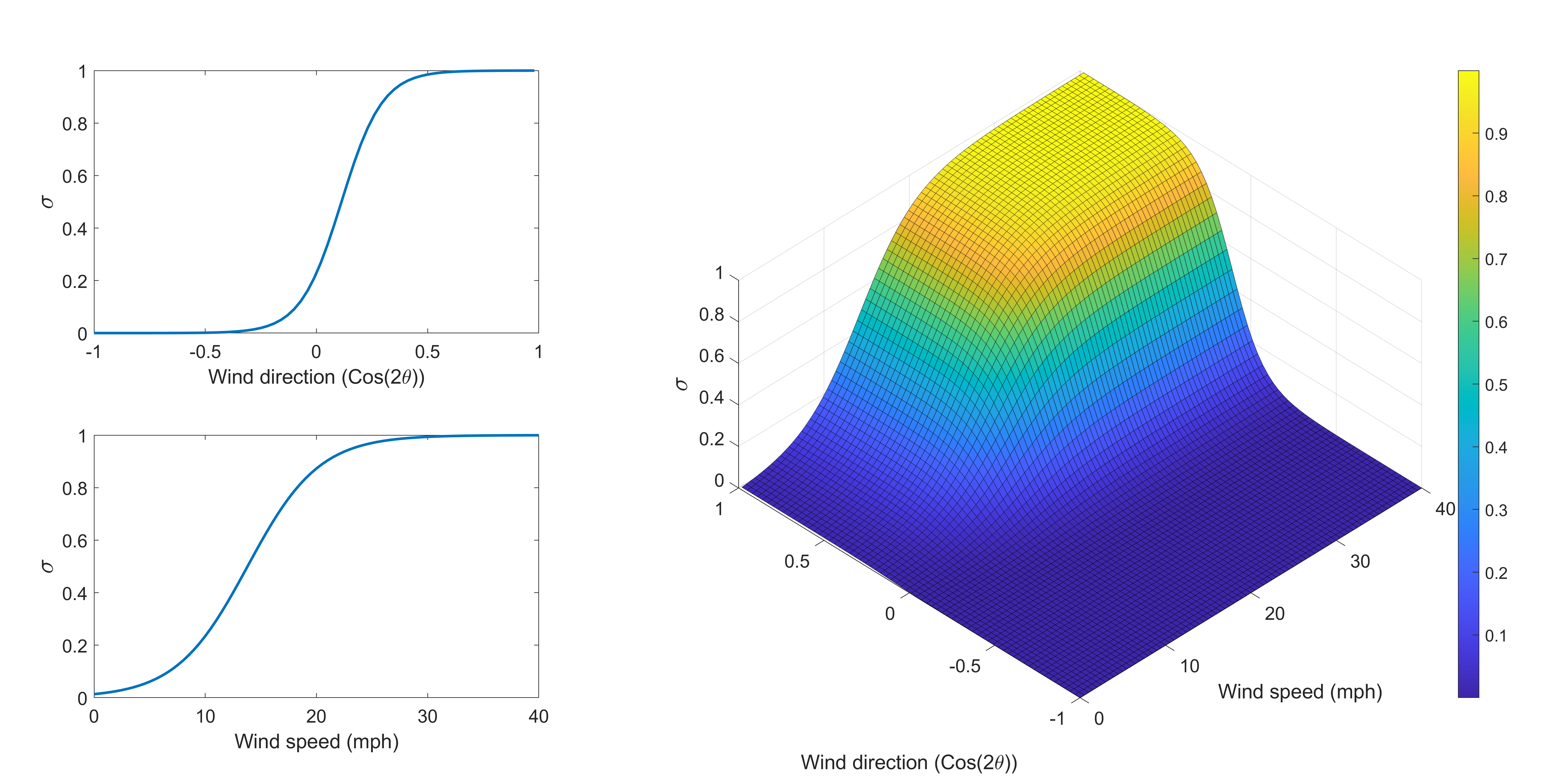}
  \caption{Learned sigmoid functions for the wind direction (top left) and wind speed (bottom left) from the physics-informed change-point kernel. The 2D sigmoid surface (right) across both wind direction and speed highlights the region in yellow in which the physics-based kernel becomes active. This represents a region of northerly/southerly winds at high speeds.}
  \label{fig:Tamar_Sigmoids_2D}
\end{figure}

Another common motivation for PIML methods is an ability to provide additional interpretability within results. This is achieved here through the physically informative hyperparameters of the sigmoid kernels. Plots displaying the sigmoids learned from the data, constructed using these hyperparameters, are shown in Figure \ref{fig:Tamar_Sigmoids_2D}. These plots are able to intuitively display, in terms of where and how quickly, how the relationships modelled within the data change. Here, the model was able to learn from the data that changing direction of wind rapidly introduced lift force as wind approached northerly and southerly directions ($\cos(2\theta) \to 1$). The dependency of the relationship on wind speed was more gradual, seen within the shallower gradient of the learned wind speed sigmoid. The yellow region of the 2d sigmoid surface highlights a region of high speed northerly and southerly winds in which the lift force kernel has the highest explanatory power within the model. Here, we had a known regime change in mind that we were trying to capture, however similar plots could be of particular help where one might be less sure about how behaviours within a dataset change.

%

\subsection{Heteroscedastic noise}

The implementation of the change point model has provided a much improved ability to capture the switch in the underlying governing physics of the deck acceleration. One final characteristic of the deck acceleration yet to be discussed is the variance of the data as wind speed and direction changes. Considering Figure \ref{fig:NESW_Subplots_data_only}, one can see that as the wind speed increases for southerly winds, the variance on the data grows. This trend is less visible for northerly winds due to the lack of data available in this direction, but would show a similar trend if higher speed northerly winds were observed in the monitoring campaign. 

For the models implemented so far, the assumption (arising from standard GP theory) is made that the observation noise (see Equation \ref{eq:tamar_obs_model}) can be modelled as a draw from a Gaussian distribution with zero mean and variance $\sigma_n^2$. More succinctly, deck acceleration $a$ is modelled as,

\begin{equation}\label{matt:eq1}
  a = f(\theta,U) + \epsilon, \hspace{5mm} f \sim \mathcal{GP}(0,K), \hspace{5mm} \epsilon \sim \mathcal{N}(0,\sigma_n^2)
\end{equation}

where the latent function $f$ captures both the and lift and nominal/operational forces through the switching kernel. Such a model provides no mechanism for allowing the variance on the measurements to change dependent upon the input values (e.g. wind speed), instead imposing a noise model with a fixed variance. This results in the uncertainty bounds that are returned by the previous models (Figures \ref{fig:NESW_Subplots_SE_Kernel}, \ref{fig:NESW_Subplots_Changepoint_Kernel}) being too large at low wind speeds in northerly and southern directions, and is a consequence of having to compensate for the larger variance seen at higher wind speeds. A more appropriate model would allow for tighter confidence regions at lower wind speeds, that can widen at higher wind speeds. 

To that end, we now consider a heteroscedastic noise process, where the observation noise itself becomes a function of the inputs, resulting in a model of the form,

\begin{equation}\label{matt:eq2}
  a = f(\theta,U) + \epsilon(\theta,U), \hspace{5mm} f \sim \mathcal{GP}(0,K), \hspace{5mm} \epsilon \sim \mathcal{N}(0,r(\theta,U))
\end{equation}

This formulation adds an additional step into the inference procedure; estimating the new observation variance term $r(\theta,U)$. A common approach is to place a second Gaussian process prior over the term, effectively treating the noise variance as a latent function to be learnt instead of a fixed parameter, and can be expressed as,

\begin{equation}
     r(\theta,U) \sim \mathcal{GP}(\mu_r,K_r).
\end{equation}

where $\mu_r$ is the mean of the noise process, and $K_r$ the covariance function. Here, we use a linear mean function to avoid imposing an arbitrary noise scale, with the squared exponential function used for the kernel. The overall model therefore now contains two GPs; one that governs the predictions of the noiseless underlying process, and another that captures the noise. Such a model is defined as a heteroscedastic Gaussian process (HGP). Although the use of a heteroscedastic noise model enhances the predictive power of the model, both the evidence and predictive distributions (the conditional mean and variance) are no longer analytically tractable. To perform inference here, we follow the variational approximation scheme as detailed in \cite{lazaro2011variational}. 

\begin{figure}[ht!]
  \centering
  \includegraphics[width=0.9\textwidth]{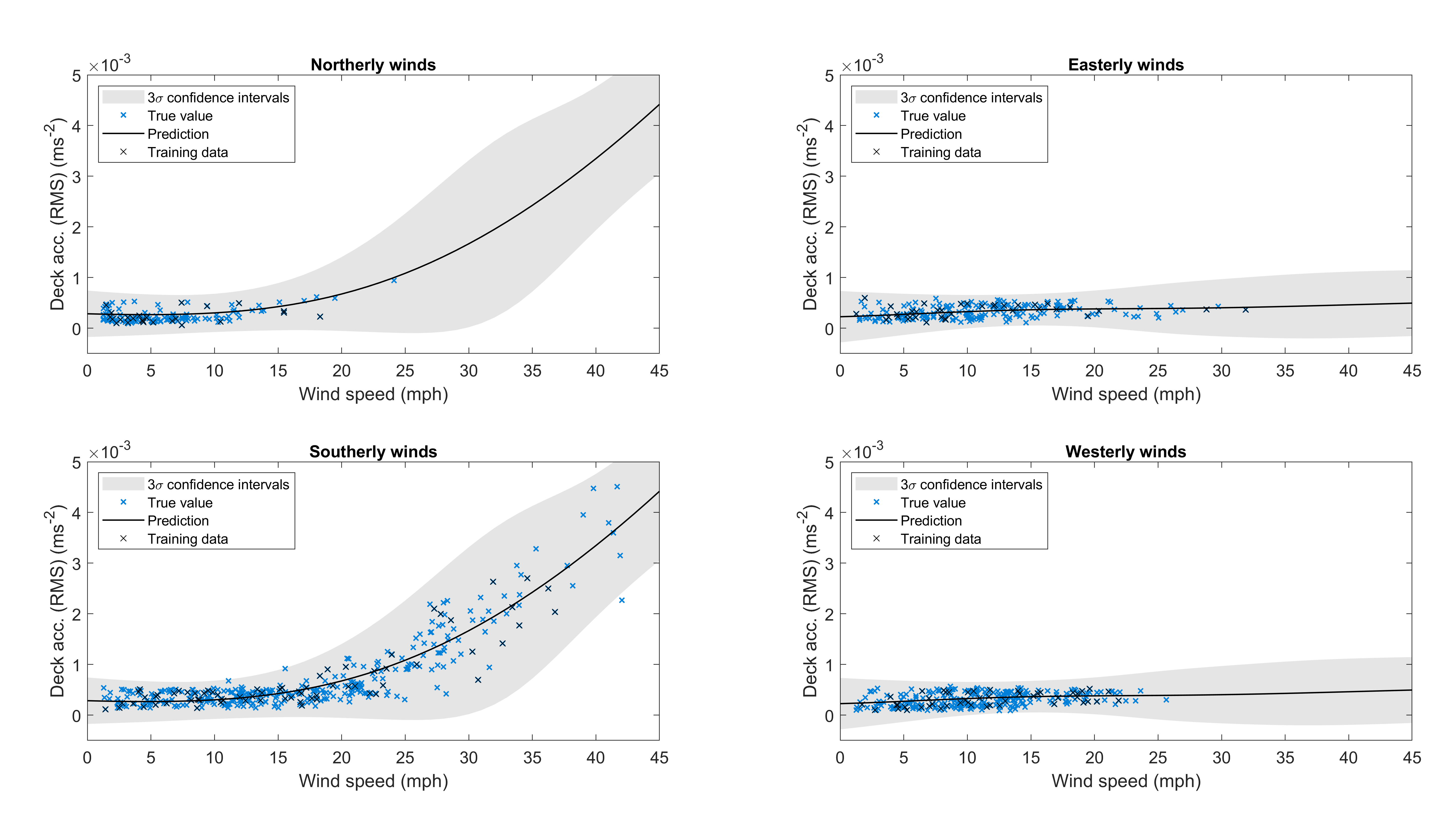}
  \caption{Predictions of vertical deck acceleration on the Tamar bridge, separated by wind direction, using a heteroscedastic Gaussian process with a physics-informed change-point kernel. The model was shown the same random scatter of 500 datapoints from a 2500 point dataset during training as the previous two models.}
  \label{fig:NESW_Subplots_HGPCP_kernel}
\end{figure}

The change-point kernel is now combined with the heteroscedastic noise process, and the model is learnt and evaluated with the same train-test split as the previous section. The results are plotted in Figure \ref{fig:NESW_Subplots_HGPCP_kernel}.

The standout observation is that the uncertainty bounds that are returned by the heteroscedastic model are now much more representative of the variation of the deck acceleration, particularly for north-south winds. We can see that at lower speed north-south winds, the predictive distributions now much more closely enclose the test data than with a standard GP noise model. The confidence bounds are then able to grow as the wind speed increases, inline with the measured data in this region. For the east-west winds, the predictive distribution returned by the heteroscedastic model also offers improved uncertainty bounds that are tighter around the test data. Considering the NMSE of the heteroscedastic model, there is little difference with that obtained by the standard change-point model. As the noise model does not appear to bias the mean predictions that are made by the standard change-point model (Figure \ref{fig:NESW_Subplots_Changepoint_Kernel}), incorporating a heteroscedastic noise model is not expected to significantly alter the mean predictions. However, considering the MSLL, a loss of -0.97. is returned, demonstrating an improvement over the standard change-point model in capturing a distribution over deck acceleration values. 



\section{Case study B: Changing physical behaviour of an aircraft wing}
Here, a case study focuses on the prediction of aircraft wing strain during an turn and how a changing physical relationship is induced due to the inputs of the pilot. Reed \cite{Reed2007TucanoStrain} previously achieved success modelling wing strain with MLP networks, resulting in an ability to estimate fatigue damage for the wing within $\pm 3\%$. Later work of Fuentes \cite{Fuentes2014InputAug} and Gibson \cite{Gibson2023DistFatigue} on the dataset used GP regression to enable access to a full predictive distribution over the strains. Gibson explored how draws from the predictive distribution may be propagated to achieve distributions over the fatigue damage.

\begin{figure}[ht]
  \centering
  \includegraphics[width=1\textwidth]{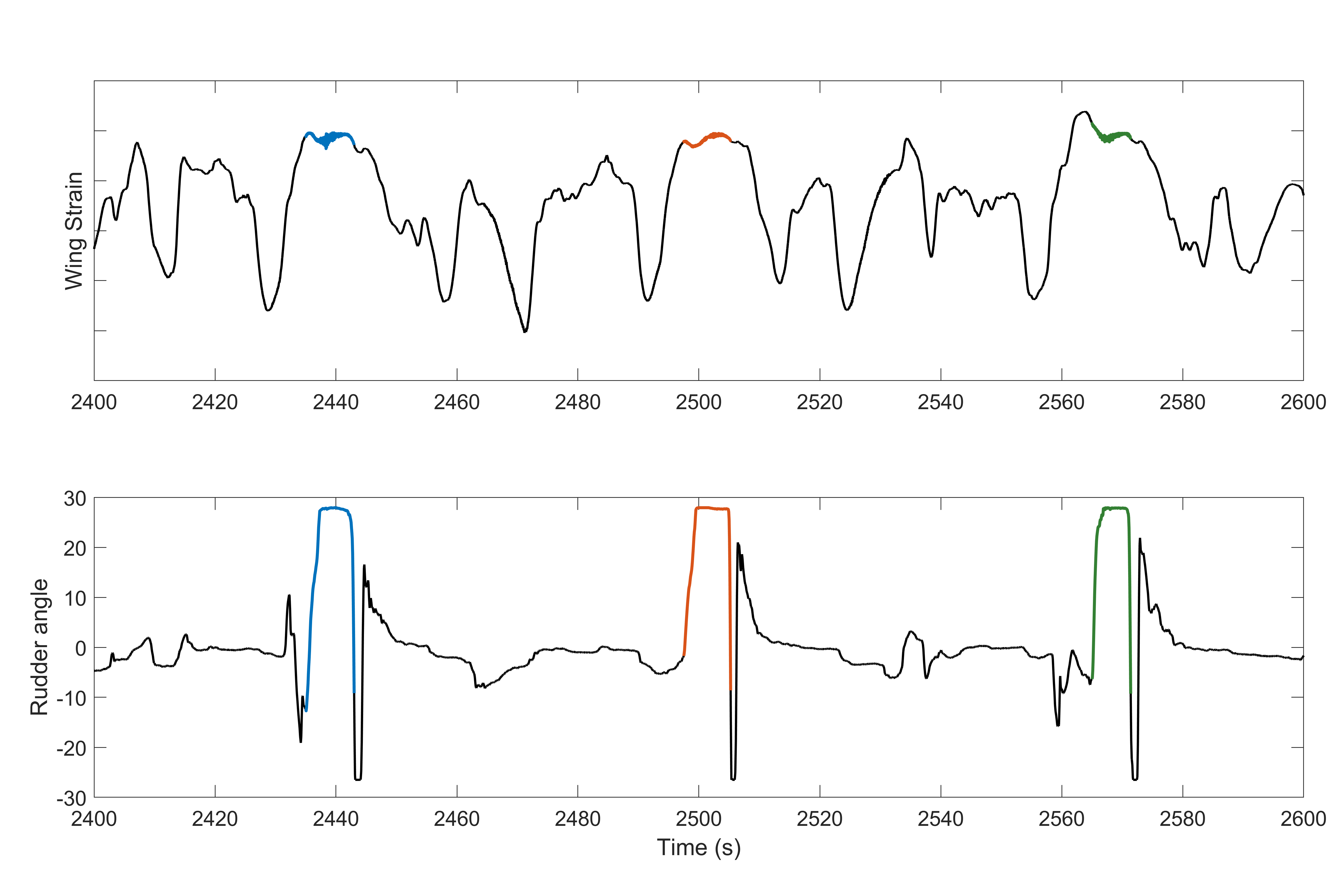}
  \caption{Time series plots of wing strain and rudder angle. Three consecutive, aerial turns are highlighted in blue, orange and green.}
  \label{fig:Tucano_3_Turn_Time_Series}
\end{figure}

\begin{figure}[ht]
  \centering
  \includegraphics[width=1\textwidth]{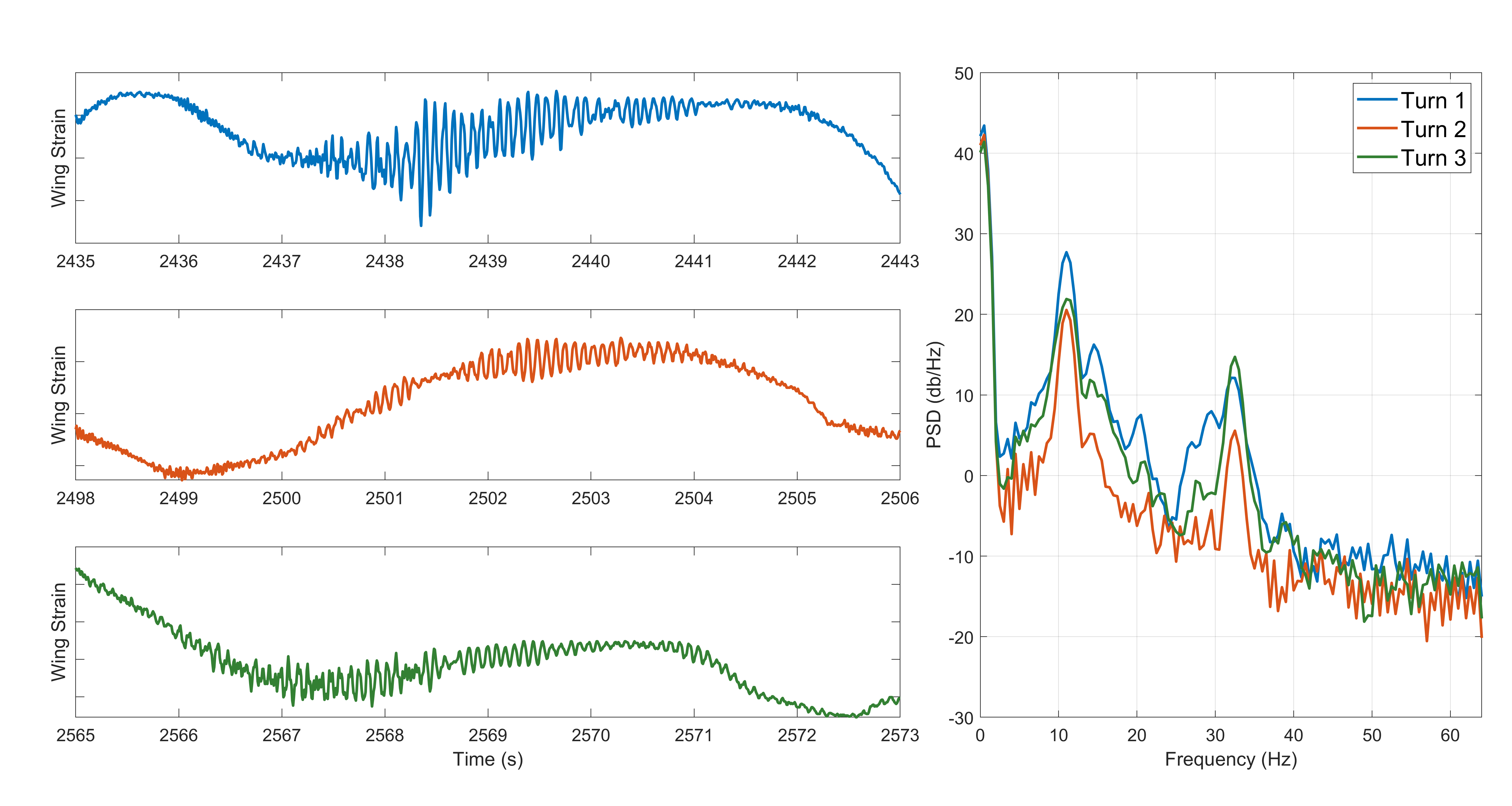}
  \caption{A comparison of the time series and frequency content of the three sharp aerial turns from Figure \ref{fig:Tucano_3_Turn_Time_Series}. The same colours: blue, orange and green are used for each turn. Two peak frequencies of $11Hz$ and $32.5Hz$ are observed across the three turns.}
  \label{fig:Tucano_3_Turn_Zoomed_P3B_With_Spectra}
\end{figure}

In this work we will illustrate the proposed method using a dataset from an aircraft in flight which includes time histories of strain measurements and pilot controls (see \cite{Reed2007TucanoStrain,Fuentes2014InputAug,Gibson2023DistFatigue}). It is worth emphasising that, where previous work focussed on the accuracy of fatigue damage estimation \cite{Reed2007TucanoStrain,Gibson2023DistFatigue}, the main aim here is the identification and modelling of changing physical phenomena. For this reason, the work predicts wing strain from a subsection of the dataset, where the pilot used an input to the rudder to cause the aircraft to turn. A time series of the wing strain and rudder angle is shown in Figure \ref{fig:Tucano_3_Turn_Time_Series}. At high rudder angles, an aircraft will turn about its $Z$ axis, and the introduction of some additional dynamic behaviour can be seen. A comparison of the induced dynamic behaviours across the three aerial manoeuvres is shown in Figure \ref{fig:Tucano_3_Turn_Zoomed_P3B_With_Spectra}. A similar frequency content, with peaks at $11Hz$ and $32.5Hz$ is shared between the three turns.


\subsection{Kernel design}
The wing strain is considered to consist of two main parts: a quasi-static component, capturing the more slowly varying relationship between flight parameters and strain; and a dynamic component, induced by oscillations occurring during manoeuvres. In this case, the manoeuvre of interest is a sharp turn due to a high rudder angle. To represent this, the problem is structured as:
\begin{equation}
y(X) = \underbrace{f_1(X)}_{\begin{matrix}{\text{Quasi-static component}}\end{matrix}} + \underbrace{f_2(X)}_{\begin{matrix}{\text{Dynamic, manoeuvre-induced behaviour}}\end{matrix}} + \underbrace{\epsilon}_{\begin{matrix}{\text{Noise}}\end{matrix}}
\end{equation}

To capture the relationship between measured flight parameters and the wing strain, inline with previous kernel selections within the literature \cite{Fuentes2014InputAug,Gibson2023DistFatigue}, we employ the Squared Exponential (SE) kernel:
\begin{equation}
	K_{SE}(X_{Fl},X_{Fl}')= \sigma_f^2 \text{exp}\left (-\frac{1}{2}(X_{Fl})^T \Lambda^{-1}(X_{Fl}')\right ) + \sigma_n^2\delta_{ij}
\end{equation}

where $\sigma_f^2$ is the signal variance, $\sigma_n^2$ is the noise variance, $\Lambda$ is the matrix of length scales such that $diag(\Lambda)=[l_1^2,l_2^2,...,l_D^2]$ for a \(D\) dimensional input and $X_{Fl}$ is a matrix of measured flight parameters.

To capture the oscillatory behaviour, highlighted in Figure \ref{fig:Tucano_3_Turn_Zoomed_P3B_With_Spectra}, a covariance function is employed that can model vibrating systems and has been directly derived from an equation of motion of a stochastic vibrating system (SDOF kernel \cite{Cross2022SDOFkernel}):
\begin{equation}
	K_{SDOF}(\tau)= \frac{\sigma^2}{4 m^2 \zeta \omega_n^3} e^{-\zeta \omega_n |\tau|} \left( cos(\omega_d \tau) + \frac{\zeta \omega_n}{\omega_d} sin(\omega_d |\tau|) \right)
\end{equation}

where the hyperparameters of the kernel now relate to physical properties of a SDOF oscillator: $m$ is the mass, $\zeta = c/2\sqrt{km}$ is the damping ratio, $\omega_n = \sqrt{k/m}$ is the natural frequency and $\omega_d = \omega_n \sqrt{1 - \zeta^2}$ is the damped natural frequency. Draws from this kernel are constrained to obey the behaviour of an SDOF oscillator, a useful property to encode. Here the SDOF kernel aims to capture the oscillatory behaviour of the wing, similarly to how previous implementations of the kernel have been used to represent modes of an oscillating cantilever beam \cite{Pitchforth2022ISMA,Jones2023IWSHM}. Due to the presence of two peak frequencies, believed to be caused by two modes of the wing being excited, it is necessary to use a sum of SDOF kernels:
\begin{equation}
  K_{MDOF} = \sum_{i=1}^{N} K^i_{SDOF}(t,t')
\end{equation}

where, here, $N=2$. A feature of the dataset, and one of the primary reasons it has been selected for this case study, is that the dynamic behaviour is intermittent. To allow the SDOF kernel to be phased in and out appropriately, it is integrated in a change-point structure of the form:
\begin{equation}
  \begin{split}
  K(X,X') & = \underbrace{K_{SE}(X_{Fl},X_{Fl}')}_{\begin{matrix}{\text{Quasi-static signal component captured with an SE kernel...}}\end{matrix}} \\
  & + \underbrace{K_{\sigma}(\text{RUDD},\text{RUDD}')\left(\sum_{i=1}^{N} K^i_{SDOF}(t,t') \right)}_{\begin{matrix}{\text{... whilst introducing the SDOF kernel(s) for rudder-based manoeuvres.}}\end{matrix}}
\end{split}
\label{eq:TucanoKernel}
\end{equation}

where $X_{Fl}$ is a matrix of flight parameters, RUDD is the rudder angle and $t$ is the time vector. Note an important difference here compared with the Tamar Bridge kernel construction in Equation \ref{eq:TamarKernel}; rather than switching to and from a pair of kernels ($K_{Lift}$ and $K_{SE}$ for the Tamar case study), the SE kernel $K_{SE}$ remains active and the physics-informed kernel $K_{SDOF}$ is being phased in and out. The most appropriate way to utilise sigmoid kernels, and therefore define how the kernel may switch, will vary by use case.

\subsection{Results}
A focus of the earlier Tamar Bridge case study was the utilisation of physical knowledge to assist with extrapolation, here, for the second case study, an emphasis is instead placed on interpolation. A key strength of the SDOF kernel is an ability to help with the upsampling of response data \cite{Cross2022SDOFkernel}. For the test cases shown here, models were shown conditions with data at 64Hz and tasked with upsampling to 128Hz. The three manoeuvres, highlighted in Figure \ref{fig:Tucano_3_Turn_Time_Series}, took place over a time period of 200 seconds, resulting in a test set size of 25,600 datapoints and a training set size of 12,800 datapoints. Without modification of the approach to a sparse GP framework \cite{Titsias2009Sparse,GPsForBigData}, one would not typically wish to adopt training set sizes beyond approximately 10,000-15,000 datapoints due to computational and memory demands.

\begin{figure}[ht]
  \centering
  \includegraphics[width=1\textwidth]{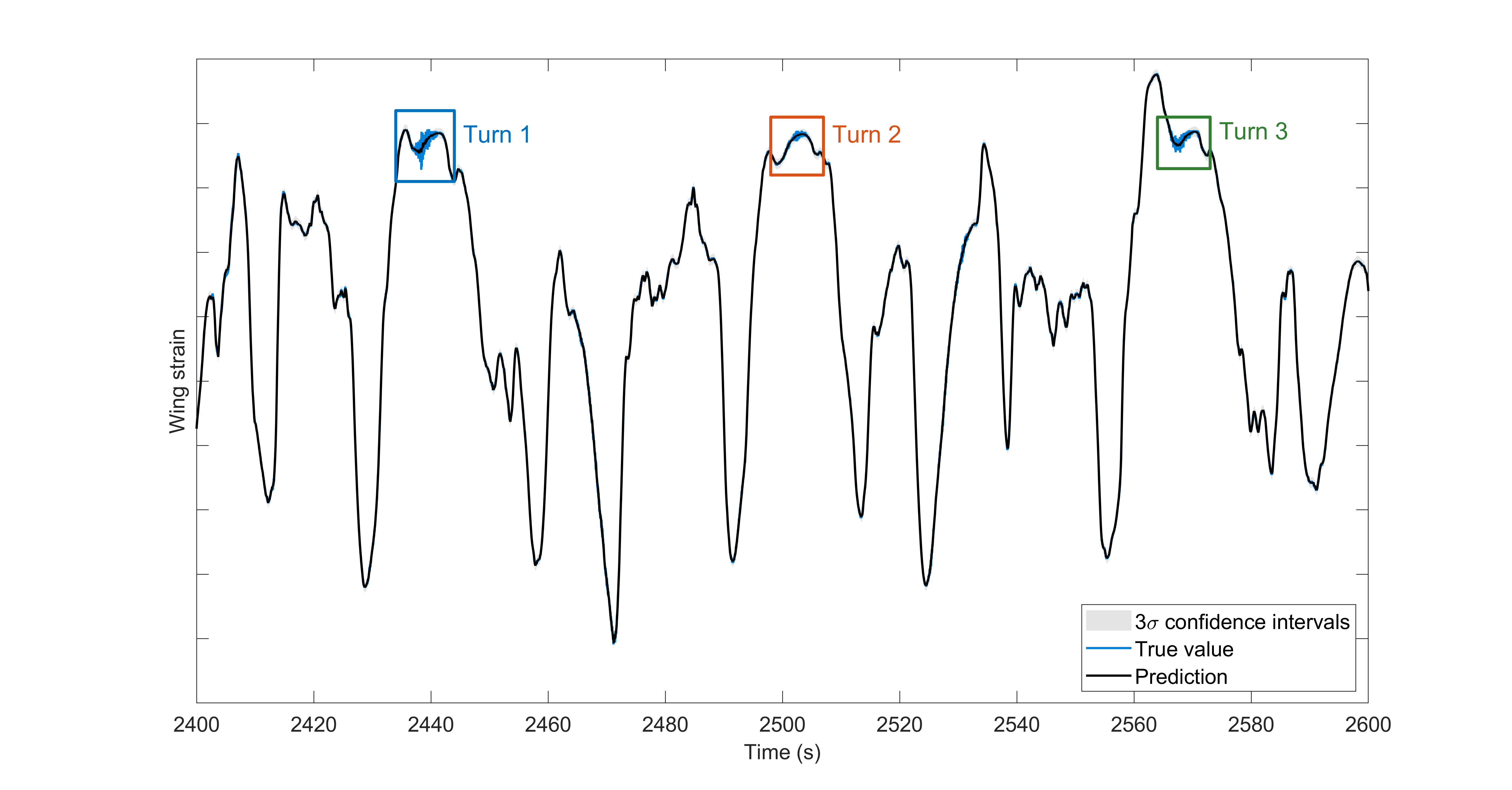}
  \caption{SE kernel prediction of inner port side wing strain (P3B) during a series of in-flight manoeuvres. The model has observed every second data point of a 128Hz, 200s time series. The model achieves a good fit for the majority of the time series, however, performance worsens within regions around a series of three sharp aerial turns.}
  \label{fig:Tucano_SE_Full_Turn_Boxes}
\end{figure}

To highlight the key areas for potential improvement, the problem was first modelled using a typical black-box approach, $\mathcal{GP}(0,K_{SE}(X_{Fl},X_{Fl}'))$: A zero mean GP, with a squared exponential kernel acting on a matrix of flight parameters, $X_{Fl}$. The timeseries prediction of this model is shown in Figure \ref{fig:Tucano_SE_Full_Turn_Boxes}. For the majority of the time series, the prediction of the SE kernel achieved a good fit to the data, with an NMSE of 0.0138\% obtained across the full time series. As a task of interpolation, one would expect a black-box model to perform well here. The model performance is worst where the dynamic behaviour is induced due to the turning of the aircraft. These regions are highlighted with blue, orange and green boxes in Figure \ref{fig:Tucano_SE_Full_Turn_Boxes} and are the primary areas of interest when comparing the change-point and SE model structures.

\begin{figure}[ht]
  \centering
  \includegraphics[width=0.8\textwidth]{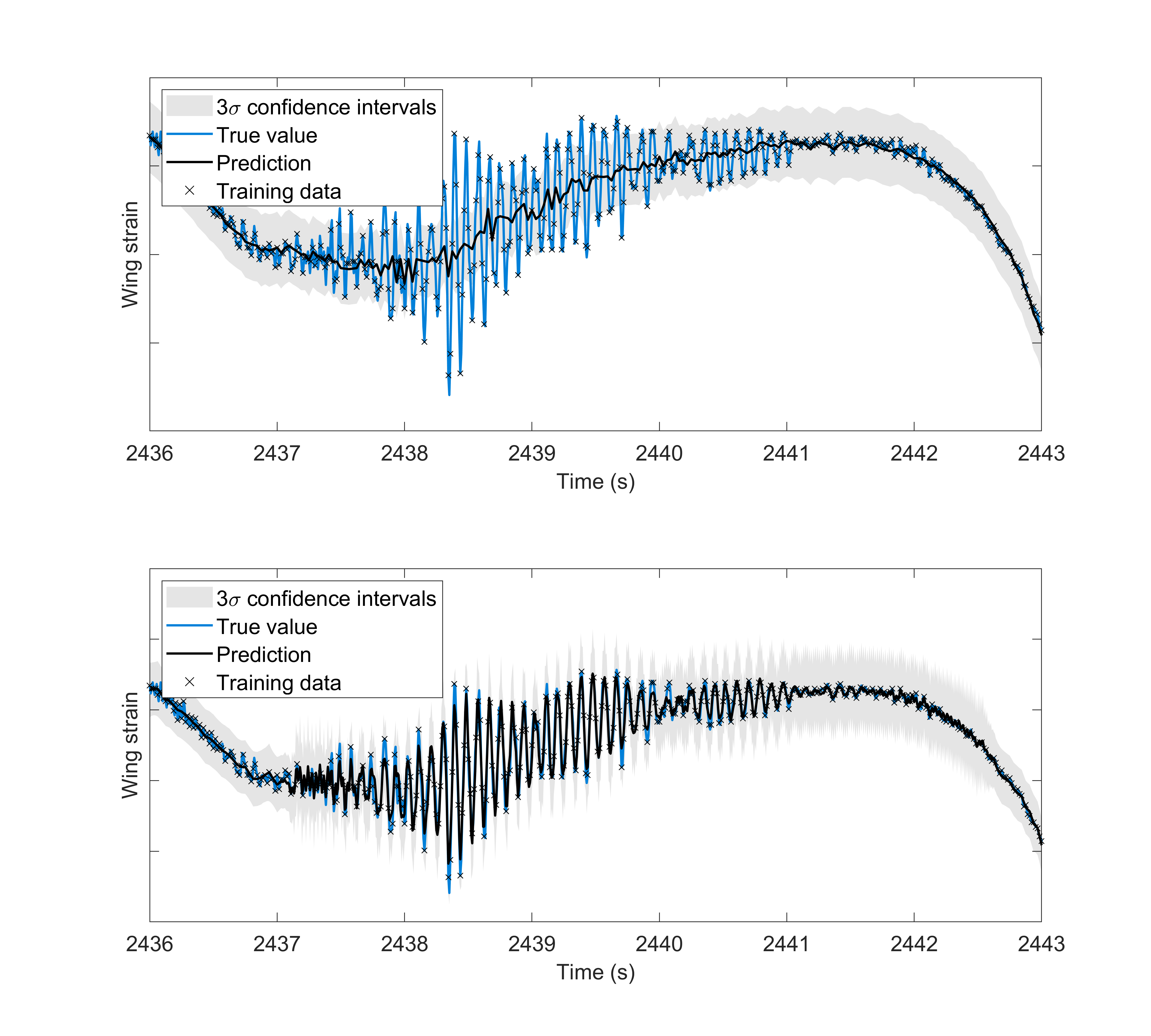}
  \caption{A comparison of SE kernel and physics-informed switching kernel (Equation \ref{eq:TucanoKernel}) predictions of inner port side wing strain during the first of three in-flight turns. The SE kernel (Top) struggles to characterise the higher frequency dynamic behaviour effectively.}
  \label{fig:Tucano_KCP_SE_Turn_1}
\end{figure}

\begin{figure}[ht]
  \centering
  \includegraphics[width=0.8\textwidth]{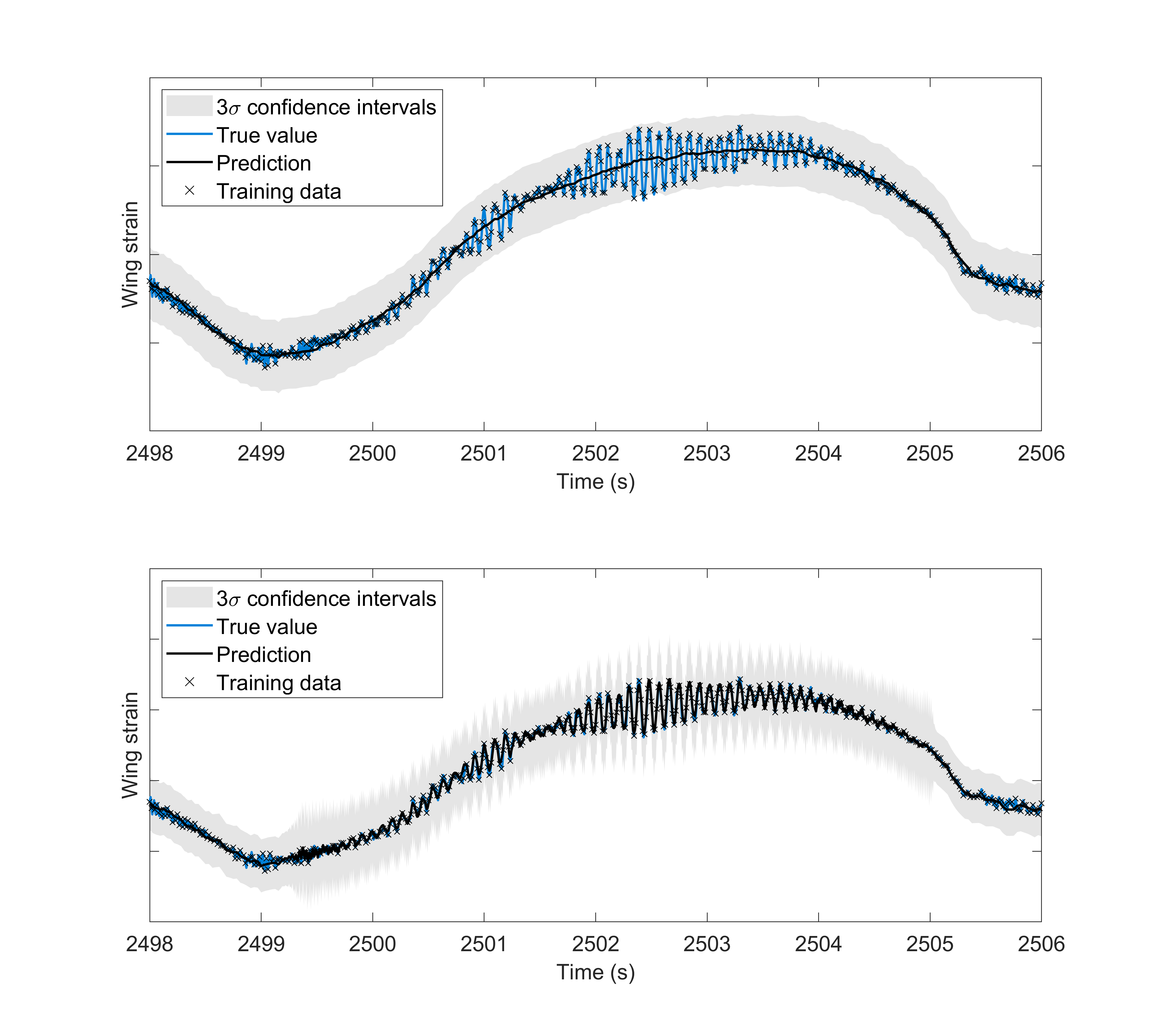}
  \caption{A comparison of SE kernel and physics-informed switching kernel (Equation \ref{eq:TucanoKernel}) predictions of inner port side wing strain during the second of three in-flight turns. The SE kernel (Top) struggles to characterise the higher frequency dynamic behaviour effectively.}
  \label{fig:Tucano_KCP_SE_Turn_2}
\end{figure}

\begin{figure}[ht]
  \centering
  \includegraphics[width=0.8\textwidth]{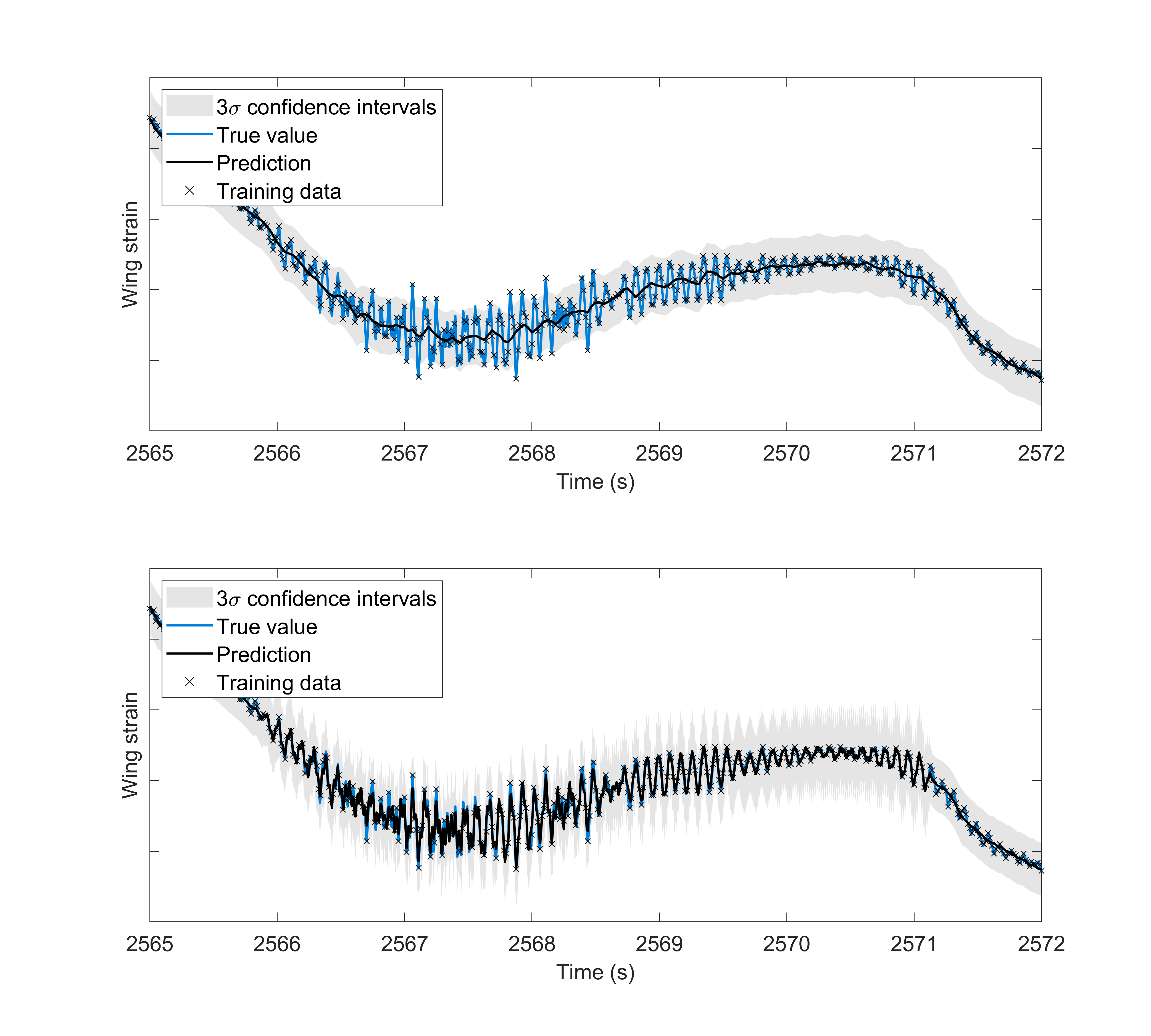}
  \caption{A comparison of SE kernel and physics-informed switching kernel (Equation \ref{eq:TucanoKernel}) predictions of inner port side wing strain during the third of three in-flight turns. The SE kernel (Top) struggles to characterise the higher frequency dynamic behaviour effectively.}
  \label{fig:Tucano_KCP_SE_Turn_3}
\end{figure}

A comparison of SE kernel and Change-point kernel performance, in terms of NMSE and MSLL, for the full timeseries and each of the three manoeuvres is shown in Table \ref{tab:TucanoPerformance}. Comparisons of time series predictions for each of the three manoeuvres are shown in Figure \ref{fig:Tucano_KCP_SE_Turn_1}, Figure \ref{fig:Tucano_KCP_SE_Turn_2} and Figure \ref{fig:Tucano_KCP_SE_Turn_3} respectively.

\vspace{0.15cm}
\begin{table}[ht]
  \sisetup{round-mode=places,round-precision=3}
    \begin{center}
      \caption{Wing strain model performance comparison.}
      \label{tab:TucanoPerformance}
      \begin{tabular}{ccccc} 
      \toprule
      \multirow{ 2}{*}{Region} & \multicolumn{2}{c}{NMSE} & \multicolumn{2}{c}{MSLL}\\
      \cmidrule(lr){2-3}\cmidrule(lr){4-5}
       & SE kernel & Change-point & SE Kernel & Change-point \\[0.2cm]
      \midrule
      Full timeseries  &  0.0138    &    0.00749   &     -4.424    &     -4.615\\ 
      Turn 1        &       22.851    &       2.944   &      0.255    &     -1.759\\     
      Turn 2        &       1.876    &      0.301   &     -1.968    &     -2.107\\    
      Turn 3       &        5.413   &      0.872    &     -1.484     &    -2.095\\
      \bottomrule
      \end{tabular}
    \end{center}
\end{table}
\vspace{0.15cm}

The performance of both models across the full time series was significantly better than during any of the three aerial-turns. For the SE kernel, the NMSE of 0.0138\% for the full time series worsened to 22.851\% during the first turn, where the model did a poor job of characterising the oscillatory behaviour. The 64Hz observed sample rate was not high enough for the SE kernel to reconstruct the higher frequency behaviour present within the 128Hz strain measurements. Instead, the model smoothed through the dynamic content, missing out on much of the structure present. As a side effect of this, a relatively high estimate was placed on the noise present within the signal, inflating the confidence bounds higher than necessary away from the dynamic content. A point of note for the SE kernel prediction of the first turn is that it achieved a positive MSLL of 0.255. An MSLL of zero represents a constant prediction of the dataset mean, with a predictive variance equal to the dataset variance. Exceeding an MSLL of zero is an indicator a model is not working correctly.

The prediction of the physics-informed change-point kernel, across the full timeseries produced an NMSE of 0.00749\%, almost halving the NMSE achieved by the black-box model. A large contributor to this was a greatly improved ability to capture the dynamic content through the phasing in of a SDOF kernel during turning manoeuvres. The NMSE for turns 1, 2 and 3 dropped from 22.851\%, 1.876\% and 5.413\% for the SE kernel to 2.944\%, 0.301\% and 0.872\%, with the physically-informed change-point kernel better able to capture the structure present within the signal. Instead of smoothing through the oscillations, the SDOF kernel was able to effectively capture the higher frequency content and provide an appropriate quantification of uncertainty. The inclusion of the SDOF kernel also allowed the SE kernel to better capture the static component of the signal. Since the SE kernel was not attempting to capture two distinct behaviours, a longer lengthscale and smaller noise variance were able to be used away from the dynamic behaviour, thereby improving overall performance.

\subsection{Model interpretability}
\begin{figure}[ht]
  \centering
  \includegraphics[width=0.8\textwidth]{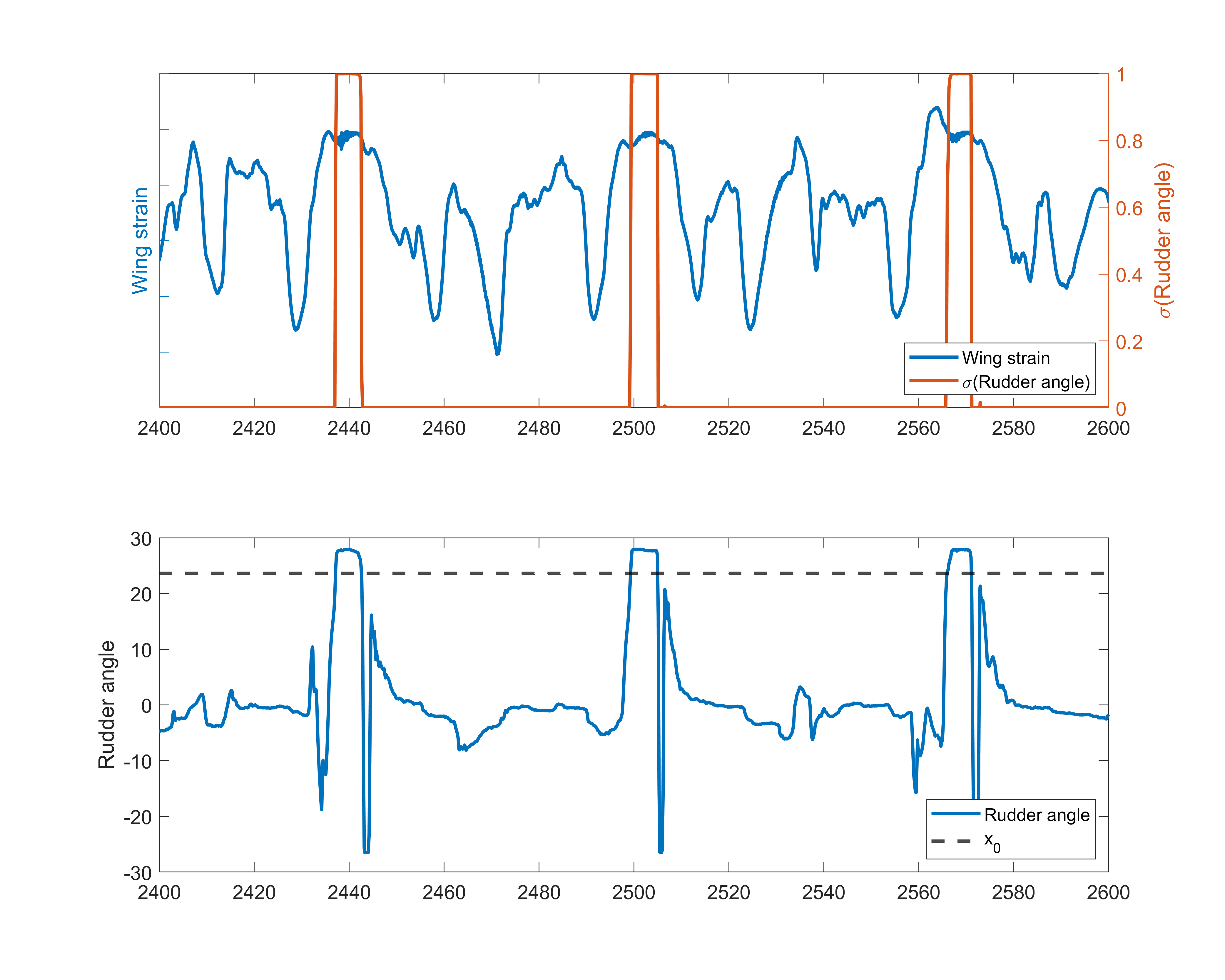}
  \caption{Interpretation of model switching using the learned sigmoid kernel hyperparameters. (Top) A time series plot of the rudder angle passed through the sigmoid function, $\sigma(\text{Rudder angle})$. The `active' region is observed to coincide with the oscillatory behaviour seen in the wing strain. (Bottom) A time series plot of the rudder angle, showing where it exceeds the sigmoid switching location $x_0$.}
  \label{fig:Tucano_Switch_Interp}
\end{figure}

As also discussed in the Tamar Bridge case study, another notable benefit from PIML models alongside increased performance is an additional degree of insight in to model results. For the change-point kernel, this came from two main sources: the hyperparameters learned from the sigmoid kernel, providing information about how the model was switching between kernel components, and the hyperparameters of the SDOF kernel, relating to physical properties of an SDOF oscillator. Here, we look to interpret the model's switching behaviour, with plots constructed using the learned sigmoid kernel hyperparameters shown in Figure \ref{fig:Tucano_Switch_Interp}.

The top subplot of Figure \ref{fig:Tucano_Switch_Interp} shows the reconstructed sigmoid function from the learned hyperparameters $x_0 = 22.59^{\circ}$ and $a = 3.099$. Within the change-point structure, this produces a switching of the SDOF kernel from fully inactive, to fully active within in a window of approximately $21-24^{\circ}$ rudder angle. The relatively steep gradient of the sigmoid, combined with a rapid change of rudder angle during the in-flight manoeuvre, produces a sudden switch within the model. An almost step function like time series plot of the rudder angle passed through the sigmoid function $\sigma(\text{RUDD})$ can be seen within the bottom subplot of Figure \ref{fig:Tucano_Switch_Interp}. For the modelling of in-flight manoeuvres, one might expect a fairly sudden switching of behaviour, however for any application, similar plots to this can provide a useful way to interpret when and how different parts of the model are being phased into or out of predictions. 

\newpage 
\section{Summary and conclusions}
This paper developped the first physics-informed model to treat the relative reliance upon physical knowledge and measured data as a switching problem. The `switch' itself was argued as a valuable source of interpretability into model operation and was captured using a novel physics-informed change-point kernel structure. The model incorporates physical knowledge within the kernel of a Gaussian process whilst being able to vary the extent to which the model relies upon this knowledge. This provides a promising tool to cope with both regime-switching and varied presence of physical behaviours. Two engineering case studies, summarised in Table \ref{tab:ChangePtSummary}, were used to highlight how the model structure can be used to improve performance and provide insight in to a models' results.

\vspace{0.15cm}
\begin{table}[ht]
  \sisetup{round-mode=places,round-precision=3}
    \begin{center}
      \caption{Summary of physics-informed change-point model implementations.}
      \label{tab:ChangePtSummary}
      \begin{tabular}{cccc} 
      \toprule
      Application & Prediction target & Switching variable(s) & Benefits of model\\[0.2cm]
        \midrule
    Tamar Bridge & Deck response & \makecell{Wind direction, \\ Wind speed} & \makecell{Separation of relationships, \\ Capability to extrapolate} \\[0.5cm]

    Manoeuvring Aircraft & Wing strain & Rudder angle & \makecell{Capture of dynamic content, \\ Enhanced interpolation}\\[0.2cm]

      \bottomrule
      \end{tabular}
    \end{center}
\end{table}
\vspace{0.15cm}

A dataset of directional wind loading of the Tamar bridge was used to highlight a case where physical knowledge of a phenomena may only apply in particular regimes. The lift force induced by high wind speed, represented here with a polynomial kernel, only acted when wind hit the bridge side-on. The change-point kernel was able to learn this and present results of the changing regimes in an interpretable manner. The model was able to extrapolate for regions of high speed northerly winds, for which a scarcity of measured data was available. Additionally, it was shown how the change-point kernel may be used in a heteroscedastic setting, where observation noise is a function of the inputs; in this case, wind speed, providing better fitted confidence bounds.

Data from an aircraft was used to investigate oscillatory behaviour within the wings, induced during in-flight manoeuvring. The behaviour was found to occur during sharp turns of the aircraft (high rudder angles), and was represented using a SDOF kernel within the change point structure. The model was able to effectively upsample from 64Hz to 128Hz and provide a breakdown of static and dynamic signal components. Compared with a purely data-based approach, model performance during three aerial manoeuvres was improved from 22.851\%, 1.876\% and 5.413\% to 2.944\%, 0.301\% and 0.872\% respectively, with performance across the complete time series improving by almost a factor of two.

\section*{Acknowledgements}
The authors gratefully acknowledge the support of The University of Sheffield and Innovate UK through the OLLGA project grant 10040817. We would like to thank the Vibration Engineering Section (VES) for their work collecting the Tamar bridge dataset.

\newpage
\bibliography{Bibliography}

\end{document}